\documentclass[12pt]{article}
\usepackage[legalpaper,  margin=1in]{geometry}
\usepackage[utf8]{inputenc}
\usepackage{bm}
\usepackage{amsmath,amsfonts,amssymb,amsthm,epsfig, graphicx, hyperref, mathtools, appendix, comment, caption, subcaption, float,setspace, booktabs}
\usepackage[flushleft]{threeparttable}
\def\supm{^{(m)}}

\usepackage[dvipsnames,table,dvipsnames*, svgnames*, hyperref]{xcolor}
\newtheorem{thm}{Theorem}

\newcommand{\beq}{\begin{equation}}
\newcommand{\eeq}{\end{equation}}
\newcommand{\beas}{\begin{align*}}
\newcommand{\eeas}{\end{align*}}
\newcommand{\bea}{\begin{align}}
\newcommand{\eea}{\end{align}}
\newcommand{\bei}{\begin{itemize}}
	\newcommand{\eei}{\end{itemize}}
\newcommand{\ben}{\begin{enumerate}}
	\newcommand{\een}{\end{enumerate}}
\newcommand{\bet}{\begin{theorem}}
	\newcommand{\eet}{\end{theorem}}
\newcommand{\bel}{\begin{lemma}}
	\newcommand{\eel}{\end{lemma}}
\newcommand{\bep}{\begin{proposition}}
	\newcommand{\eep}{\end{proposition}}
\newcommand{\bed}{\begin{definition}}
	\newcommand{\eed}{\end{definition}}
\newcommand{\bec}{\begin{corollary}}
	\newcommand{\eec}{\end{corollary}}
\newcommand{\bex}{\begin{example}}
	\newcommand{\eex}{\end{example}}

\definecolor{red}{RGB}{200,50,150}
\definecolor{darkred}{RGB}{150,50,50}
\definecolor{brown}{RGB}{250,100,100}
\definecolor{green}{RGB}{000,150,100}
\definecolor{purple}{RGB}{100,000,250}

\def\blue{\color{blue}}
\def\red{\color{red}}

\def\red{\color{red}}

\def\blue{\color{blue}}

\newcommand{\bu}{\bold{u}}

\newcommand{\bw}{\bold{w}}

\newcommand{\bbf}{\bold{f}}

\newcommand{\bD}{\bold{D}}

\newcommand{\balpha}{\boldsymbol{\alpha}}

\newcommand{\bbeta}{\boldsymbol{\beta}}
\newcommand{\bmu}{\boldsymbol{\mu}}
\newcommand{\btheta}{\boldsymbol{\theta}}

\newcommand{\bz}{\bm{z}}

\usepackage[utf8]{inputenc}
\usepackage{amsmath}
\usepackage{amsfonts}
\usepackage{amssymb}
\usepackage{algorithm}
\usepackage{algpseudocode}



\usepackage{natbib}
\usepackage{multirow}
\usepackage[utf8]{inputenc}
\def\D{\mathbf{D}}

\newcommand{\YY}{\boldsymbol{Y}}
\newcommand{\xx}{\boldsymbol{x}}
\newcommand{\uu}{\boldsymbol{u}}
\newcommand{\yy}{\boldsymbol{y}}

\newcommand{\R}{\mathbb{R}}
\newcommand{\E}{\mathbb{E}}

\newcommand{\argmin}{\mathop{\rm arg\min}}
\newcommand{\argmax}{\mathop{\rm arg\max}}

\newcommand{\vertiii}[1]{{\left\vert\kern-0.25ex\left\vert\kern-0.25ex\left\vert    1 
		\right\vert\kern-0.25ex\right\vert\kern-0.25ex\right\vert}}

\newcommand{\nb}[1]{{\bf\blue    1}}
\newcommand{\nr}[1]{{\bf\red    1}}

\newcommand{\eqnr}[1]{{\beq {\red    1}\eeq}}
\newcommand{\eqr}[1]{{\[ {\red    1}\]}}
\newcommand{\indep}{\rotatebox[origin=c]{90}{$\models$}}

\def\xx{\bold{x}}
\newtheorem{remark}{Remark}
\title{Federated One-Shot Ensemble Clustering}
\author{Rui Duan$^{1,\dag}$, Xin Xiong$^{1}$, Jueyi Liu$^{2}$, Katherine P. Liao$^{3,4}$, Tianxi Cai$^{1,3,\dag }$\\ \\
    1 Department of Biostatistics, Harvard T.H. Chan School of Public Health \\ 
    2 Department of Biostatistics, University of Michigan \\ 
    3 Department of Biomedical Informatics, Harvard Medical School \\ 
    4 Division of Rheumatology, Inflammation, and Immunity,\\ Brigham and Women’s Hospital \\ 
    $\dag$ Co-corresponding authors}

\date{}

\begin{document}

\maketitle
\doublespacing 

\begin{abstract}
      Cluster analysis across multiple institutions poses significant challenges due to data-sharing restrictions. To overcome these limitations, we introduce the Federated One-shot Ensemble Clustering (FONT) algorithm, a novel solution tailored for multi-site analyses under such constraints. FONT requires only a single round of communication between sites and ensures privacy by exchanging only fitted model parameters and class labels. The algorithm combines locally fitted clustering models into a data-adaptive ensemble, making it broadly applicable to various clustering techniques and robust to differences in cluster proportions across sites. Our theoretical analysis validates the effectiveness of the data-adaptive weights learned by FONT, and simulation studies demonstrate its superior performance compared to existing benchmark methods. We applied FONT to identify subgroups of patients with rheumatoid arthritis across two health systems, revealing improved consistency of patient clusters across sites, while locally fitted clusters proved less transferable. FONT is particularly well-suited for real-world applications with stringent communication and privacy constraints, offering a scalable and practical solution for multi-site clustering.\\
      
\end{abstract}

\section{Introduction}

Clustering analysis is a core task in data science that involves grouping data points into clusters according to their similarity \citep{hagenaars2002applied}. As the cluster labels are not predetermined \citep{celebi2016unsupervised}, this process is considered a form of unsupervised learning.  Clustering analysis methods can generally be divided into two main types: mixture models, which assume that data points are generated from a mixture of probability distributions, and model-free methods, which focus on specific measures of similarity or dissimilarity \citep{mclachlan2019finite,mclachlan1988mixture,ng2001spectral}. In biomedical research, clustering analysis has many applications, including identifying subtypes of disease which are meaningful for disease management and treatment design \citep{fereshtehnejad2017clinical,grisanti2022neurological}, inferring population structures in genetic and genomic research \citep{jakobsson2007clumpp,han2017clustering}, and  differentiating between different types of tissue or cell types based on imagaing or gene expressing data \citep{kiselev2019challenges}.

With the growing opportunities for multi-institutional collaborative research,  joint analyses across multiple sites are oftentimes preferred over separate, study-specific analyses, due to the improved sample sizes, interpretability, and diversity of the overall study population which potentially improves the  generalizability of the results.  However, conducting multi-site joint clustering analysis poses significant challenges under data sharing constraints, particularly when individual-level data cannot be shared across sites \citep{malin2013biomedical,guinney2018alternative}. 
In such settings, clustering methods which require assessing the similarity or dissimilarity among data points are difficult to be implemented across sites since similarity between two data points from two different studies cannot be directly evaluated. On the other hand,  locally identified clusters at individual sites may lack direct comparability due to label switching, particularly in cases of large estimation errors, and widely-used evidence synthesis methods based on averaging over sites may not be feasible. In addition to restrictions on sharing individual-level data,  the constraint on communication costs is another practical consideration. It's important to minimize both the amount of information required from each site and the frequency of communications between site to ensure real-world applicability. To address the above mentioned challenges, we propose a {\bf F}ederated {\bf O}ne-shot e{\bf N}semble clus{\bf T}ering (FONT) algorithm which requires only one round of communication across study sites, and it also protects against sharing raw individual-level data.

Existing federated clustering methods can be categorized into several types. One type of approaches involves identifying landmark data points within each site locally and then grouping these landmarks into clusters \citep{magdalinos2006k,chen2011large}. 
For instance,  the K-fed method \citep{11} identifies local landmarks as cluster centers at each site, and obtains global clusters by applying the K-means algorithm on these local landmarks. Although K-fed is a one-shot algorithm, it shares common limitations with other landmark-based methods: first, landmarks are typically defined based on a certain distance measure, making them less effective for handling various types of data such as longitudinal, categorical, or time-to-event data. Second, poor data quality or limited sample size at local sites can result in unreliable landmarks, with errors that cannot be corrected during subsequent landmark clustering.
Many other federated clustering methods require iterative communications across sites for updating model parameters or cluster memberships \citep{10,8,4,qiao2024federated}. For example, some methods are based on a federated optimization procedure, often implemented by applying divide-and-conquer to the gradient descent steps \citep{3,8}. Additionally, \cite{7} used soft-clustering, which iteratively calculates cluster centers and memberships until convergence. However, iterative procedures typically require data-sharing infrastructures across sites, which may not be available in many collaborative research environments.

In this paper, we propose the FONT algorithm as a general federated clustering framework based on a data-adaptive ensemble of locally fitted clustering models. This method is communication-efficient, requiring only a single round of communications across sites, and protects individual's privacy by only sharing fitted model parameters and class labels across sites. Our approach is applicable to a wide range of clustering techniques, from non-parametric algorithms like K-means to parametric latent class models. It accommodates varying cluster proportions across sites and is robust enough to manage scenarios where certain clusters may be absent at some sites. The FONT algorithm is related to the consensus clustering methods \citep{strehl2002cluster, monti2003consensus, li2007solving,xanthopoulos2014review}, where the key idea is to combine multiple partitions to yield a consensus clustering. Some methods focus on combining clustering results obtained by re-sampling from the original data \cite{monti2003consensus}, others are more generally focusing on using methods such as similarity-based or graph-based partitioning \cite{strehl2002cluster}, non-negative matrix factorization \citep{li2007solving}, K-means clustering on a combined similarity matrix obtained from the cluster memberships to obtain a combined clustering. Therefore, most of the consensus clustering methods give equal weights to all clusterings, which is less efficient and less robust to settings where some of the clustering results are poor. \cite{li2008weighted} proposed a weighted consensus clustering method allowing assigning different weights to different clusterings. However, their algorithm does not have a principled way to identify the weights, and there is no guarantee on the effectiveness of the selected weights. Our proposed FONT algorithm distinguishes itself from the existing consensus clustering methods mostly from two aspects: We construct distance matrices for all individuals using the estimated model parameters, which more accurately capture individual-level similarity compared to relying solely on cluster memberships; we provide theoretical guarantees for the data-adaptive weights, demonstrating that these weights align with the accuracy of the local models.

The following sections of the paper is organized as follows. In Section \ref{sec:method}, we introduce in details the problem setting and the proposed FONT algorithm. We provide theoretical justifications on the choice of the data-adaptive weights. In Section \ref{sec:simulation}, we validate the  effectiveness of the FONT algorithm through simulation studies in various settings. In Section \ref{sec:data}, we applied the FONT algorithm to train a latent class model using sequences of medication records extracted from two health systems. Our study identified four latent subgroups among patients with rheumatoid arthritis.  A brief summary and discussion on  future directions based on the limitation and limitations, practical considerations can be found in Section \ref{sec:discussion}.
%\begin{enumerate}
%         \item One-shot, communication efficient, only cluster membership is shared, no variable information is required
 %        \item Applicable to a broad class of clustering methods, including K-means,  or parametric latent class model. 
 %        \item Performance is stable to challenging scenarios including outliers, missing clusters in certain sites.
%     \end{enumerate}

%\section{Literature review-TBD}

\section{Method}\label{sec:method}

\subsection{Problem set-up}

Suppose there are a total of $N$ subjects from $M$ study sites, where each subject's data are only stored at one of the $M$ sites. For the $i$-th individual,  we use $R_i \in [M] \equiv \{1,...,M\}$ to denote the site indicator with $R_i = m$ indicating that the $i$-th individual is in the $m$-th site. We denote the $p$-dimensional observed variables to be $X_i\in \mathcal{X}\subseteq\R^p$.  We assume there are $K$ latent classes where each individual belongs to one of the $K$ latent classes, and the unobserved class label is denoted by $Y_i\in [K]$.

We assume that across sites, each cluster is defined consistently, i.e.,  $X\indep R|Y$. To account for cross-site heterogeneity, we assume that the cluster proportions can vary across sites, that is, $p_{km} = P(Y = k|R = m)$ may not be the same as $p_{km'} =P(Y = k|R = m')$. We assume that $K$ is fixed and the overall class proportion $p_k=P(Y = k)>\delta>0$ for some constant $\delta$. %{\red although $p_{km}$ can be 0 for some site $m$.} 
%\tcomm{Rui confirm?} 
%\dcomm{Yes}

We consider a class of clustering methods with the clustering rule indexed by $\btheta\in \mathbb{R}^d$, i.e., $h_{\btheta}:\mathcal{X}\rightarrow[K]$, where $d$ grows with $K$ and $p$, and can potentially grow with the sample size $N$. %\tcomm{clarify if the dimension of $\btheta$ can grow.}\dcomm{It can grow, but we do not have explicit rate of convergence requirement on the local clustering results. When $d$ is large the signal to noise condition for the ensemble learning may not hold} 
  When $\btheta$ is given, the cluster membership for any data point  can be inferred from $h_{\btheta}(\xx)$. We further assume that  $\btheta$ includes cluster specific parameters $\bbeta_1,\bbeta_2,\dots,\bbeta_K$, where each of them corresponds to each cluster although the cluster labels $[K]$ may be assigned arbitrarily. Besides the cluster specific parameters, there might be additional parameters indicated by $\balpha$. For example, in $\balpha$ can be a feature representation projecting the original features to a lower dimensional space. This class of methods include mixture models, where the conditional distribution can be specified as $
X|{Y=k} \sim f(\xx, \bbeta_k).
$ 
%and let $\btheta = ( \btheta_1, \dots, \btheta_K)$, which is shared across all sites, i.e., $X\perp R|Y$. %To account for data heterogeneity, we assume that the cluster proportions can vary across sites, that is, $p_{km} = P(Y = k|R = m)$ may not be the same as $p_{km'} =P(Y = k|R = m')$. %To be more general, we can also allow the probability of being in one
Given $\{\bbeta_k\}_{k = 1}^{K}$, the class label of a subject with observed data $\xx$ from site $m$ can then be predicted by
\[
\hat y =\argmax_k 
P(Y = k|X = \xx, R=m) = \frac{p_{km} f(\xx; \bbeta_k)}{\sum_{s\in[K]} p_{sm} f(\xx; \bbeta_s)}.
\]
In addition to mixture models, the class of  methods also extend to non-parametric clustering approaches, provided these methods can determine the cluster membership of new data points using only summary statistics, allowing for the classification of new data without needing to access the original dataset used in the model's initial fitting phase. For example,  in the K-means algorithm, the cluster membership of data point $\xx$ is obtained through comparing distances between  $\xx$ to the $K$ cluster means $\{\bbeta_k\}_{\{1\le k \le K\}}$, i.e.,
\beq\label{kmeans}
\hat y (\btheta,\xx) = \argmin_{k\in\{1, \dots, K\}} \|\xx - \bbeta_k\|_2^2.
\eeq
Hence, with estimates of the $K$ cluster means, the cluster membership of any data point can be predicted. As a contrast, our method is not compatible with clustering approaches that require re-running the model or accessing individual-level information of all subjects to assign cluster membership for a new data point, such as DBSCAN \citep{schubert2017dbscan}. From a model implementation perspective, these methods face inherent limitations even in a single-site setting, as one would want to predict cluster memberships for new coming data without the need for refitting.

%Thus differences characterizes the heterogeneity of data across different sites. 

%With the estimator of the membership matrix, we will then apply appropriate clustering algorithm to estimate the class membership. 

%In a non-federated setting, similarity matrices are the foundations of many clustering algorithms. For example,  in spectral clustering algorithms, one needs to construct the similarity matrix, where the similarity between the $i$-th and $j$-th individual is often calculated as $s_{ij} = x_i^Tx_j$, or $s_{ij} = \exp(-\|x_i-x_j\|_2^2/h)$ with some bandwidth $h$. In a federated setting  the similarity matrix cannot be calculated as $x_i$ cannot be shared directly to a different site. 

%For example, if assuming a Gaussian mixture model with identity, $\theta$ can be the cluster
%The clustering algorithm is restricted to a class where the output of the algorithm $\mathcal{A}$ includes the estimates of a set of parameter $\theta\in \R^d$, with which we can predicted the class membership of a new data point $x$ as a function of the parameter $\theta$ and $x$, i.e, $\hat{y} = h(\theta,x)$. 
%of a new data point $x\in \R^p$, and given the value of some finite dimensional parameter $\theta\in \R^d$, denoted as $\hat{y} = \mathcal{A} (\theta,x)$. 

%More importantly, we assume that given $\theta$,  we predict the class membership of any data point $x$ by a function of $\theta$ i.e, $\hat{y} = h(\theta,x)$. 

When all data can be pooled together, we can fit a clustering model directly to obtain class labels for all $N$ subjects. In a federated setting with data sharing constraints, challenges arise due to difficulties in assessing similarity between subjects at different sites, label switching issues, and data heterogeneity affecting the global performance of locally trained models. To address these challenges, we propose the FONT algorithm, as detailed in Section 2.2, that uses an ensemble learning strategy to learn a proxy for the true similarity/dissimilarity among subjects, assigning adaptive weights to locally fitted models to account for varying data quality and noise levels.

\subsection{FONT algorithm}

%In this seTo overcome the aforementioned challenges, we propose measure the distance (or similarity) among the $N$ subjects based on summary statistics. 

Given the challenges in assessing distances among individuals using original data due to data sharing constraints, we propose representing each subject by the parameter corresponding to their latent class. We define a symmetric distance matrix $\D=[D_{ij}]_{N\times N}$, where 
\beq\label{distance}
D_{ij} = d(\sum_{k = 1}^{K}\mathbb{I}(y_i = k)\bbeta_k,\sum_{k = 1}^{K}\mathbb{I}(y_j = k)\bbeta_k)
\eeq
and $d(\cdot)$ is a pre-specified distance function such as the Euclidean distance. The between-subject distance is essentially measured by the corresponding interclass distance. The distance $\D$ is fully informative for identifying the cluster memberships and hence considered as the oracle distance matrix. Throughout the section we assume that  the number of clusters $K$ in the overall population is known. Without prior knowledge about $K$, we provide empirical methods of choosing $K$  in {Remark \ref{remark_chooseK}}.

In the federated setting, we can fit clustering models locally at each site and obtain  estimators $\{\hat \btheta^{(m)}\}$ for $m\in [M]$.  These estimates are summary-level statistics which can be shared across sites. However, these estimators are subject to label switching such that $\hat \bbeta^{(m)}_k$ and $\hat \bbeta^{(m')}_k$ may not referring to the same latent class so they are not directly comparable.  With each $\hat \btheta^{(m)}$ and {for each $1\le i\le N$,} we can obtain an estimated class label  $\hat y_i^{(m)}$ of the $i$-th subject estimated from the $m$-th model {based on the clustering rule $h_{\hat \btheta^{(m)}}$.}  The class label  $\hat y_i^{(m)}$ does not contain information regarding the original data $X_i$, and it can be shared with a central analytical server along with the pseudo identification number $i$, or a summary table reporting the fitted sample sizes of each  cluster. We denote the predicted class membership vector of all $N$ subjects to be $\hat{\yy}^{(m)} \in [K]^N$. 

With membership predicted by the $m$-th local model, we construct the distance matrix $\hat \D\supm=[\hat D\supm_{ij}]$ with
\beq\label{Dm}
\hat D^{(m)}_{ij}= d\bigg(\sum_{k = 1}^{K}\mathbb{I}(\hat y_i^{(m)} = k)\hat\bbeta_k^{(m)},\sum_{k = 1}^{K}\mathbb{I}(\hat y_j^{(m)} = k)\hat\bbeta_k^{(m)}\bigg),
\eeq
which is invariant to label switching. 
We next aggregate $\{\hat\D\supm: m\in [M]\}$ into an ensemble distance matrix 
\beq 
\widetilde \D = \sum_{m = 1}^{M} w_m\frac{\hat \D^{(m)}}{\|\hat \D^{(m)}\|_F},
\quad w_m\ge 0, \quad \sum_{m = 1}^{M}w_m^2 = 1,
\eeq
to better approximate the oracle distance matrix $\D$. We will then apply distance-based clustering algorithms (e.g., the K means algorithm) to obtain the final estimates of the cluster memberships of all $N$ subjects.

We employ a spectral ensemble  method to choose the weights $\{w_m:m\in[M]\}$ reflecting the relative quality of the clustering models fitted at the different sites. Specifically, we obtain an agreement matrix $G$ which captures the agreement among the $M$ distance matrices  $\{\hat\D\supm: m\in [M]\}$. For notational ease, we define $\text{vec}(\cdot)$ as the operator that converts a $N \times N$ matrix into a $N^2$-dimensional vector by concatenating its columns, and $\text{vec}^{-1}(\cdot)$ as the inverse operator that converts a $N^2$-dimensional vector back into a $N \times N$ matrix. For the models learned from the $t$-th and $s$-th sites, we define their agreement by
\beq\label{eq:G}
G_{ts}= \frac{\langle \bbf^{(t)},\bbf^{(s)}\rangle}{\|\bbf^{(t)}\|_2\|\bbf^{(s)}\|_2}
\eeq
and thus
\[
G = \bar F\bar F^\top\in\R^{M\times M},
\]
where $\bbf^{(m)} = \text{vec}(\hat \D^{(m)})\in \R^{N^2}$, $\bar F\in\R^{M\times N^2}$ has its $i$th row as $\bbf^{(i)}/\|\bbf^{(i)}\|_2$.  
We denote the first eigenvector of the symmetric matrix $G$ by $\widehat\bu_1=(\widehat u_{11},...,\widehat u_{1M})\in\R^M$, %Following similar ideas proposed in \cite{parisi2014ranking}, 
from which we define the weight vector $\widehat\bw=(\widehat w_1,...,\widehat w_M)$ by $\widehat w_m = |\widehat u_{1m}|$, $m=1,2,...,M$.
The final weighted distance is defined as 
\beq \label{sw}
\widetilde \D= \text{vec}^{-1}(\widetilde \bbf):=\text{vec}^{-1}(\bar F^\top \widehat\bw) =\text{vec}^{-1}\bigg(  \sum_{m = 1}^M \frac{\widehat w_m \bbf^{(m)}}{\|\bbf^{(m)}\|_2}\bigg). %\hat D^{(m)}
\eeq
We summarize the detailed steps in Algorithm 1. 
Intuitively, this approach can be considered as carrying out a principal component (PC) analysis on the matrix $\bar F$ that combines the standardized intercluster distances estimated from the $M$ sites, where $\widehat\bu_1$ is the first PC loadings of $\bar F$, under which the linear combination $\bar F^\top \widehat\bw$ of $\frac{\bbf^{(m)}}{\|\bbf^{(m)}\|_2}$  has the largest variation so as to characterize the consensus information in $\bar F$. In particular, under suitable conditions  on the overall quality of the site-specific intercluster distance estimators (see Theorem \ref{thm.main} below), the first PC loadings will have the same sign (e.g., all nonnegative) with high probability. Whenever $\{\bbf^{(m)}\}_{m =1}^{M}$ altogether contain sufficient amount of information about the underlying true intercluster distance $\bbf_0 = \text{vec}(\D)\in \R^{N^2}$, it can be shown that the components of $\widehat\bw$ would reflect the true similarity between each $\bbf^{(m)}$ and $\bbf_0$ (Theorem \ref{thm.main} Part 1), justifying our spectral weighting scheme  (\ref{sw}). Moreover, we can show that the  consensus  estimator $\widetilde\bbf$ is asymptotically no worse than the best candidate estimator among all sites (Theorem \ref{thm.main} Part 2). Similar spectral weighting idea has been considered for combining multiple classifiers without labeled data \citep{parisi2014ranking}. For our federated clustering problem, one would expect to obtain improved  clustering of all samples by applying standard methods such as hierarchical clustering \citep{murtagh2012algorithms} or K-means algorithm \citep{ikotun2023k} to the weighted consensus distance matrix $\widetilde \D$, compared with local models. 

{
\begin{remark}\label{remark_chooseK}
As common in the clustering literature, we assume that $K$, the number of
clusters, is known. In practice, for model-based approaches, many recommend using information criteria such as Akaike's Information Criterion (AIC) or Bayesian Information Criterion (BIC), although the consistency in selecting the number of classes has been investigated primarily through numerical methods \citep{mclachlan2014number}. For simpler cases such as the Gaussian mixture model, methods such as cross-validation \citep{wang2010consistent}, eigenvalue based heuristics \citep{von2007tutorial} are also available. In the FONT algorithm, for each site we can initially estimate the number of clusters locally using an appropriate method. The overall number of clusters for the population is then determined by majority voting, which works well in settings where most local sites  observe all $K$ clusters. Nevertheless, we note that even in cases where some sites experience imbalanced or missing clusters, errors arising from inaccurate estimation of cluster-specific parameters or over-specification of cluster memberships can still be mitigated through ensemble aggregation (see Theorem \ref{thm.adv}).
\end{remark}}

\begin{algorithm}[H]
\textbf{Input} Multi-institutional data $X$ of sample size $N$ stored at $M$ sites; clustering method $h_{\btheta}$; number of clusters $K$.\\
\textbf{Output} Estimated cluster membership $\tilde\yy\in[K]^N$ 
\caption{Federated ensemble learning}
\label{alg: fedc}
\begin{algorithmic}[1] 
\For{$m = 1, 2, \dots$ M}
\State Fit clustering analysis within each site and obtain parameter estimates $\hat \btheta^{(m)}$
    \State Share $\hat \btheta^{(m)}$ with all sites and the analytical center.
\State Apply clustering function $h_{\hat \btheta^{(m)}}$ to data all sites and obtain cluster membership $\hat\yy^{(m)}$
    \State Share $\hat\yy^{(m)}$ with the analytical center.
    \EndFor
\For{$m = 1, 2, \dots$ M}
\State Obtain $\hat \D^{(m)}$ according to equation (\ref{Dm}) and $\bbf^{(m)} = \text{vec}(\hat \D^{(m)})$
\EndFor
\State Obtain the 
agreement matrix $G$ according to Equation (\ref{eq:G}).
\State Apply principal component analysis on $G$ and obtain the first eigenvector $\widehat\bu_1=(\widehat u_{11},...,\widehat u_{1M})\in\R^M$, and the weight vector $\widehat\bw=(\widehat w_1,...,\widehat w_M)$ by $\widehat w_m = |\widehat u_{1m}|$, $m=1,2,...,M$.
\State Obtain the ensemble weighted distance as 
\[
\widetilde \D =\text{vec}^{-1}\bigg(  \sum_{m = 1}^M \frac{\widehat w_m \bbf^{(m)}}{\|\bbf^{(m)}\|_2}\bigg). %\hat D^{(m)}
\]
\State Apply clustering analysis on $\widetilde \D$ and obtain cluster membership $\tilde\yy$.
\end{algorithmic}
\end{algorithm}

\subsection{Theoretical justification of the data-adaptive weights}\label{sec:theory}

%\tcomm{i think we should add a collary or remarks for the recovery of the clustering membership. if you can recover $\bbf_0$, then you have the recovery of the oracle distance matrix, hence the clustering membership}

Below we provide theoretical justification for the spectral ensemble method. We start by introducing some useful notation. We denote
\[
\bbf^{(m)} = \bbf_0+\bz_m,\qquad 1\le m\le M,
\]
where $\bbf_0 = \text{vec} (\D)$ is the vectorized true distance matrix, and $\bz_m$ is the estimation error of $\bbf^{(m)}$, which we assume for now to be sub-Gaussian with variance proxy parameter $\sigma^2\equiv\sigma^2_N$, which may depend on (grow with) $N$. A more challenging noise setting is considered later in Theorem \ref{thm.adv}. We also define 
\beq
\bw=(w_1,...,w_M) = \left(\frac{\langle\bbf_0,\bbf^{(1)}\rangle}{\|\bbf_0\|_2\|\bbf^{(1)}\|_2},\frac{\langle\bbf_0,\bbf^{(2)}\rangle}{\|\bbf_0\|_2\|\bbf^{(2)}\|_2},...,\frac{\langle\bbf_0,\bbf^{(M)}\rangle}{\|\bbf_0\|_2\|\bbf^{(M)}\|_2}\right)
\eeq
characterizing the true similarity between the candidate estimators $\bbf^{(m)}$ and the true distance $\bbf_0$. Intuitively, a larger value of $w_j$ suggests that the $j$-th model provides more accurate estimations of the parameters and better identification of cluster memberships. %\tcomm{is there any optimality of $\bw$? does it minimize any uncertainty measure for any special case?}
%\dcomm{We cannot show that our weighting is the best, but we can show under what conditions it is better than averaging. And we can show that when the signal-to-noise criteria do not hold, there is no weight better than ours.} 
As will be seen shortly, a weighting scheme based on the consistent estimator of $\bw$ can lead to strictly better clustering as compared with any local model.
For any $j,k\in\{1,2,...,M\}$, we define the cross-covariance matrix $\Omega_{jk}=\E \bz_j\bz_k^\top$, and quantify the overall dependence in the errors between a pair of candidate estimators by $\rho_{jk}=\|\Omega_{jk}\|/\sigma^2$, where $\|\Omega_{jk}\|$ is the operator norm of $\Omega_{jk}$. We  denote $\rho=\|{\bf \Lambda}\|$ where ${\bf \Lambda}=(\rho_{jk})_{1\le j,k\le M}$, which quantifies the overall dependence in the errors across all local methods.  We also define the sequence $\tau_N$, which may depend on $N$, such that $\|\bz_m\|_2^2=c\sigma^2 \tau_N^2(1+o_P(1))$ for some constant $c>0$. Intuitively, the parameter $\rho$ characterizes the similarity (or overall dependence) in the errors across all the local models, whereas $\tau_N$ quantifies the average size of the errors across the local models. The ratio $\|\bbf_0\|_2/\sigma$ characterizes the overall signal-to-noise ratio across the $M$ sites.

%\tcomm{$\{\bz_m\}_{1\le m\le M}$ identically across $(i,j)$ or across $m$? elements of $\bz_m$ cannot be independent within site? also could vary by $m$? $\bw$ is also not really population parameter since it involves $\bbf^{(m)}$ which contains estimation error?}

\begin{thm}\label{thm.main}
Suppose that $\{\bz_m\}_{1\le m\le M}$ are identically distributed sub-Gaussian vectors with parameter $\sigma^2$,  and for each $m$, we have $\|\bz_m\|_2^2=c\sigma^2 \tau_N^2(1+o_P(1))$ for some constant $c>0$ and sequence $\tau_N$.  Suppose that $\|\bbf_0\|_2/\sigma\gg \log N+\tau_N\sqrt{\rho/M}$, $\rho=o(M)$,  and $M=o(N^2)$ as $N\to\infty$. Then, as $M, N\to\infty$,
\begin{enumerate}
\item For the spectral weights $\widehat\bw$ and the consensus distance estimator $\widetilde\bbf$, it holds that 
\[
\cos\angle(\widehat\bw,\bw)\to 1,\qquad \cos\angle(\widetilde\bbf,\bbf_0)\to 1,
\]
in probability.
\item For any constant $\delta\in(0,1)$, there exist a constant $C>0$ such that, whenever $\|\bbf_0\|_2/\sigma\le C\tau_N$, we have 
\[
\max_{1\le m\le M}\cos (\bbf^{(m)}, \bbf_0)<1-\delta,
\]
in probability.
\end{enumerate}
\end{thm}

Theorem \ref{thm.main} does not require independence among the $\bz_m$'s, which aligns with the practical context of our application. The magnitude of $\bz_m$ is influenced by both the classification accuracy of individual models and the estimation accuracy of interclass distances. While these factors may vary across sites, by considering an additional layer of randomness at the site level, $\bz_m$'s can be treated as realizations from some underlying distribution.

Part 1 of the theorem implies that the proposed spectral weights $\widehat\bw$ well approximates the similarity between individual site estimate of the distance matrix and the consensus distance matrix. Moreover, our result confirms the statistical consistency of the proposed consensus distance estimator $\widetilde\bbf$. {The condition ${\|\bbf_0\|_2}/\sigma\gg \log N+\tau_N\sqrt{\rho/M}$ suggests that the algorithm may benefit from a stronger overall signal-to-noise ratio, a smaller correlation among the error coordinates (i.e., smaller $\sigma\tau_N$), a larger number ($M$) of sites,  or a list of more diverse candidate estimators (i.e., smaller $\rho$) in the sense that different estimator may contain some unique information about the underlying true clusters.}

Part 2 of the theorem ensures that the proposed distance estimator $\widetilde\bbf$ is in general no worse than the individual candidate estimators, and is strictly better than the best candidate estimator under the  signal-to-noise ratio regime $\|\bbf_0\|_2/\sigma\le C \tau_N$, that is, when the overall signal strength in the $M$ sites is relatively weak. In the latter case, while each candidate estimator only has limited performance,  by Part 1, the proposed estimator is still consistent whenever $\log N=O(\tau_N)$.

Finally, we demonstrate the robustness of the proposed method in the presence of exceptionally poor local models. Specifically, we consider a more challenging setting where among the $M$ local models, only $(1-\eta)M$ of them are well-conditioned, satisfying the conditions of Theorem \ref{thm.main}, whereas the other $\eta M$ local models are misleading, which do not contain much information about the true clusters.

\begin{thm} \label{thm.adv}
Suppose among $\{\bbf^{(i)}\}_{1\le i\le M}$, there is a collection $\mathcal{C}_0$ of $(1-\eta)M$ elements for some small constant $\eta\in(0,1)$ satisfying the conditions of Theorem \ref{thm.main}, and a collection $\mathcal{C}_1$ of $\eta M$ noninformative local models such that $\bbf_0^\top\bbf^{(i)}=0$ for all $i\in \mathcal{C}_1$. Then, we still have $\cos\angle (\widetilde\bbf, \bbf_0)\to 1$ in probability. 
\end{thm}

By Theorem \ref{thm.adv}, even when there are a small portion of poor local models producing orthogonal cluster information, to be combined with the other relatively better local models, the proposed method still provides  consistent weight estimator by Part 1 of Theorem \ref{thm.main}. In other words, the poor local models will be automatically assigned nearly zero weights, to ensure the quality of the final ensemble weighted distance $\widetilde\D$. In contrast, for the naive average distance with non-informative (equal) weights across all local models, the final ensemble distance matrix will necessarily suffer from the impact due to the poor local models, rendering strict sub-optimal performance in the final clustering.

\begin{remark}
A straightforward consequence of the above theorems is that, no matter which clustering method is adopted to obtain the final cluster membership base on the ensemble weighted distance $\widetilde \bD$, it will asymptotically recover the cluster membership as if the same clustering method is applied to the oracle distance $\bD$. In particular, as long as the clustering algorithm, considered as a map from a distance matrix to a cluster membership matrix, is scale-invariant and displays certain level of continuity around the true distance $\D$, which is likely the case in many applications, Theorem \ref{thm.main} ensures that the final cluster membership will be consistently recovered from the ensemble weighted distance $\widetilde \bD$.
\end{remark}

\section{Simulation Study}\label{sec:simulation}

In this section, we generate simulation studies to investigate the numerical performance, comparing the proposed method to a few benchmark methods. We generate data from the following Gaussian mixture models in this numerical experiment: 
\beq\label{gaussianmixture}
X|Y = k, R= m \sim N(\bmu_k, \sigma\mathbf{I}_p),
\eeq
although our method is broadly applicable to many data generating mechanisms and clustering methods. 
We  consider the proportions of individuals from each latent class to be different across sites, i.e., $p_{km}$ varies over $m$. Each site also has unequal sample sizes to mimic the practical setting where some some sites can be much larger than others.

The  $K$-means algorithm is used for obtaining the local models. Each site shares its fitted cluster means $\hat\bmu^{(m)} = (\hat\bmu_1^{(m)}, \dots,\hat\bmu_K^{(m)})$ with all the  sites and  the estimated cluster membership $\hat Y^{(m)}_i$ is defined in Equation (\ref{kmeans}), which is then shared with the analysis center to obtain the aggregated distance matrix $\widetilde \bD$. We apply $K$-means again on $\widetilde \bD$ to obtain the final clusters.

\subsection{Comparison with benchmark methods}
We compare the proposed method with the following benchmark methods: 
\begin{enumerate}
    \item (local) a local K-means algorithm which do not borrow information from other sites. This is the case when we directly port a local algorithm to all sites. 
    \item (consensus) the consensus clustering method proposed by \cite{li2007solving}. The method first obtain the connectivity matrices $\{\hat S^{(m)}\}_{m=1}^{M}$ defined as 
    \begin{equation*}
\hat S^{(m)}_{ij} =\begin{cases}
1 \quad \quad \quad \textup{ if }  \hat Y^{(m)}_i =  \hat Y^{(m)}_j\\
0  \quad \quad \textup{ otherwise } \\
\end{cases}   
\end{equation*}
where $Y^{(m)}_i$ is the predicted latent class  memebership from the $m$-th local model. 
Then it obtains  the average connectivity 
    \[
    \bar S = \frac{1}{M}\sum_{m=1}^{M}\hat S^{(m)}.
    \]
    Instead of using the non-negative matrix factorization approach proposed by \cite{li2007solving} to obtain the clustering results based on $\bar S$, it has been shown that an equivalent way is to apply $K$-means algorithm to $\bar U$ where $\bar U$ satisfies $\bar S = \bar U\bar U^T$ and can be obtained by eigen decomposition of $\bar S$ \citep{li2008weighted}.
    \item (K-fed) the K-fed algorithm proposed by \cite{11}.
\end{enumerate}
In addition to these three methods which are all feasible non-iterative algorithms to obtain global clustering, we also consider two methods which are not feasible in real application scenarios. 
\begin{enumerate}
\setcounter{enumi}{4}
    \item (pooled) global clustering obtained by applying $K$-means algorithm to the pooled dataset combining M local datasets. This is infeasible in the federated setting. 
    \item (best\_local) Among all the $M$ local models, we choose the one with the best clustering accuracy. This is infeasible in practice because the true latent class membership is unknown. 
\end{enumerate}
In simulation study, we consider the pooled and the best local model as oracle methods to help us understand the potential gaps due to data sharing constraints or unknown labels. 
%Once we obtain the aggregated distance matrix using the proposed method, we apply $K$-means on it to obtain the f

Across various settings which are described in the following subsection, we fix the number of latent classes to $K=5$, and the dimension of the data $p=10$. We use multiple metrics when comparing the performance of the above mentioned methods. We first evaluate the performance of clustering accuracy,  using the adjusted rand index which measures the clustering accuracy adjusting for label switching and chance. 
We then evaluate the estimated weights $\{w_m\}_{m=1}^{M}$ in terms of how good the weights align with the estimation accuracy between each $\hat \D^{(m)}$ and the true distance matrix $\D$.

\begin{figure}
    \centering
    \includegraphics[scale = 0.5]{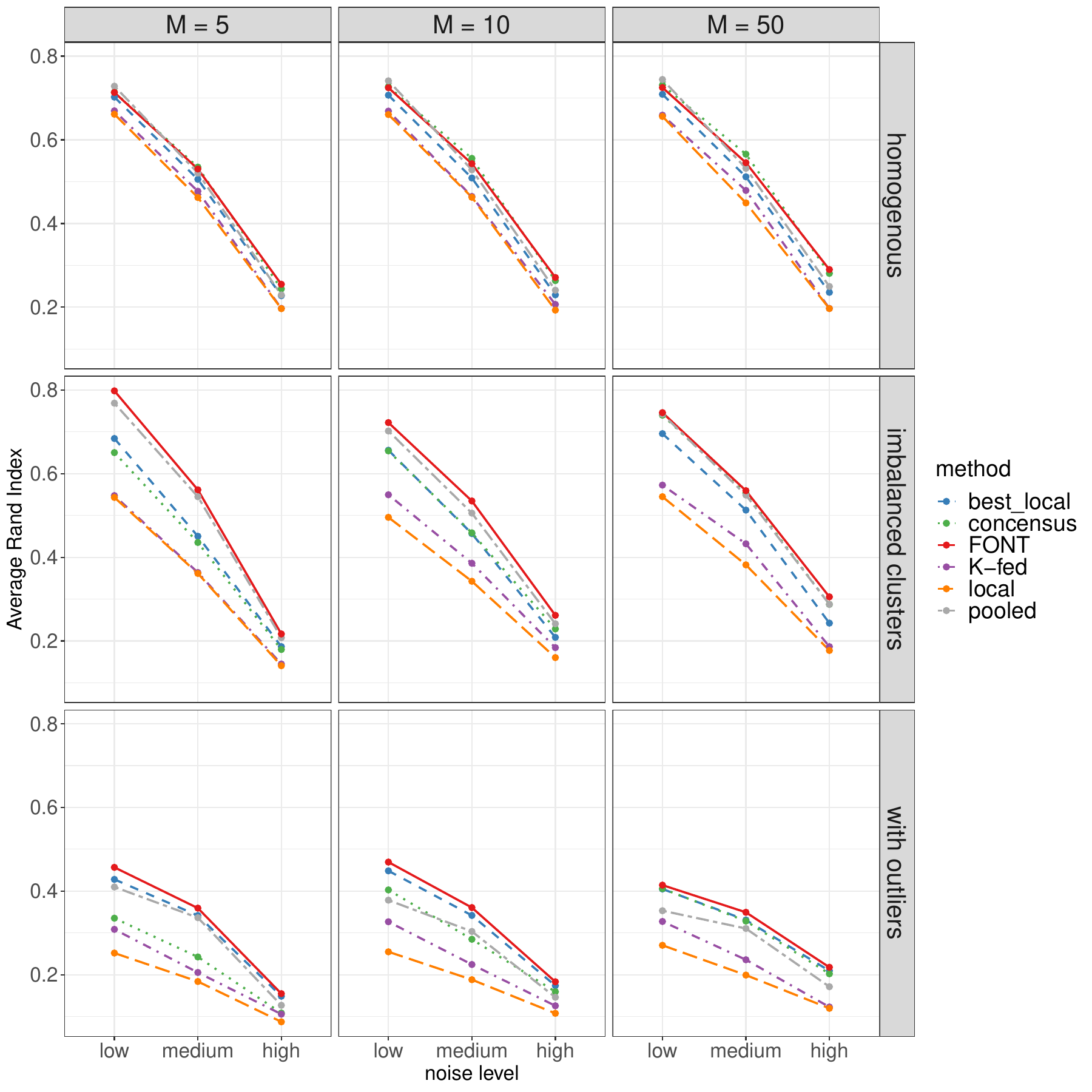}
    \caption{Model performance across different simulation settings evaluated by the average rand index. %\tcomm{change Nsite to M in all figures, change proposed to FONT}
    }
    \label{fig:allsetting}
\end{figure}

We consider the following settings with different levels and sources of heterogeneity. 

\begin{itemize}
    \item \textbf{Homogeneous setting}. In this setting, we generate independently and identically distributed data for $M=5, 10, \text{ and } 50$ sites following the Gaussian mixture model in (\ref{gaussianmixture}). We generate each entry of $\bmu_k$ randomly from -1 and 1. The noise level $\sigma^2$  takes values $0.05, 0.1, 0.3$, corresponding to the cases with high, medium and low signal to noise ratios.  The total sample size at each site ranges from $50$ to $500$. Cluster proportions are set to equal. 
    \item \textbf{Heterogeneous setting with imbalanced clusters}. In this setting,  the heterogeneity comes from the proportions of each latent classes in each site, we generate randomly from a $K$-dimensional Dirichlet distribution with parameters $\alpha_k =1$, for all $k$. Other specifications are the same as the homogenous setting. 
    \item \textbf{Heterogeneous setting with outlier contamination}.
    In this setting, in addition to the imbalanced cluster proportions, we randomly select $20$\% of the $M$ sites, within which we add outlying data with sample size $20$\% of the original data. The outlying data are generated from Gaussian distributions with every entry of the mean randomly generated from from -5 and 5. The outlying data are not assigned to any clusters and therefore are not used when evaluating the performance of the model.
\end{itemize}

Figure \ref{fig:allsetting} shows the model performance evaluated based on clustering accuracy measured by the average adjusted rand index \citep{steinley2004properties}, where  higher values indicate better performance. 

In the homogeneous setting where data distributions are the same across sites and with equal cluster proportions, we see that FONT has nearly the same as classification accuracy as the consensus clustering algorithm, which assigns the same weight to all local models. The two oracle methods, the pooled clustering analysis and the best local model,  also yield similar performance as the FONT algorithm. The local pooled analysis has similar performance as the K-fed algorithm, which are slightly lower than FONT, indicating the improvement achieved by co-clustering across sites.

In heterogeneous settings with imbalanced cluster proportions, the FONT algorithm shows a significant improvement over other approaches, particularly when the number of sites is small and the noise level is low. When $M$ increase to 50, we see that the consensus clustering has similar performance as FONT. The pooled analysis slightly outperforms the best local model in this case, since the pool data have sufficient observations from all clusters while the local data by chance might all be imbalanced. Numerically, we still see some improvement comparing FONT to the pooled method, but this might be due to the fact that ensemble learning in general can improve stability against numerical issues. In the heterogeneous setting with outlier contamination, we see similar patterns as compared to the imbalanced clusters.

\begin{figure}
    \centering
    \includegraphics[scale = 0.55]{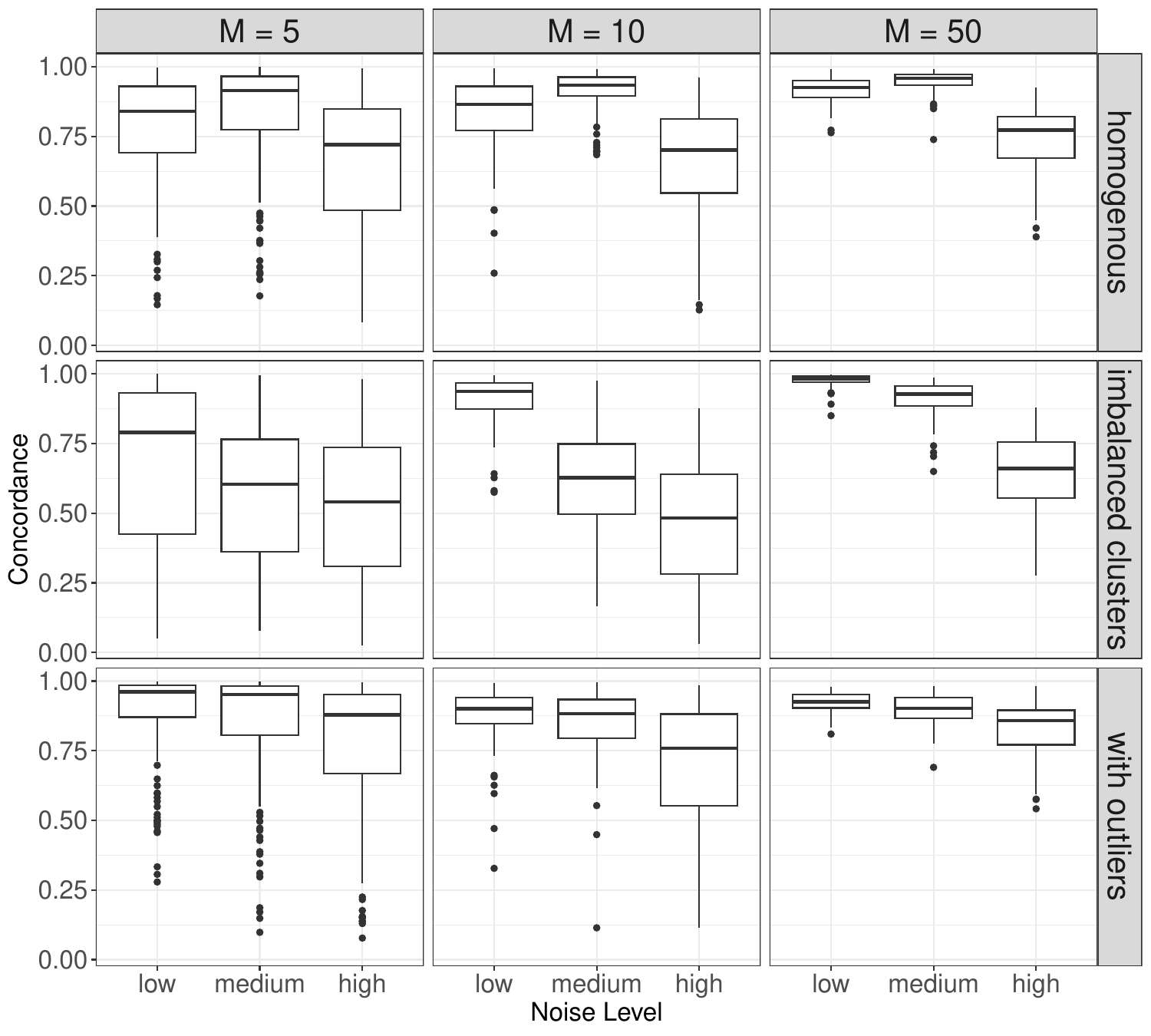}
    \caption{Correlation between the data adaptive weights received by the local models and the performance of local models evaluated by adjusted rand index.}
    \label{fig:allsetting2}
\end{figure}

\subsection{Evaluation of the data-adaptive weights}
To evaluate the weights assigned to each site, we also calculated the correlation between the weight $\{w_m: m\in[M]\}$ received by each site and the performance of the local model trained at each site. Figure 2 shows the boxplots of the correlations obtained from the all simulation settings. In most of the settings, we see that their concordance is higher with low noise level and decreases as the noise level increases, and the concordance is higher when the number of sites grows.  This aligns with the results we obtained from Part 1 of  Theorem \ref{thm.main}, where $\cos\angle(\widehat\bw,\bw)\to 1$.

%\subsection{Performance under misspecified  number of clusters}

\section{Application to latent class analysis using medication sequence data}\label{sec:data}

Over the past decade, advancements in biologic disease modifying anti-rheumatic drugs (bDMARDS) and targeted synthetic DMARDs (tsDMARDs) have broadened the treatment options for rheumatoid arthritis (RA). However, identifying the most effective DMARD for individual patients remains a significant research focus. Most studies have concentrated on phenotyping patients based on their response to tumor necrosis factor inhibitors (TNFi), the most prevalent biologic DMARD. Currently, there are five main classes of DMARDs for treating RA, targeting distinct pathways: TNFi, CTLA4, interleukin-6R (IL6R), janus kinase (JAK) inhibitors, and antibodies to CD20 (anti-CD20), comprising a total of 12 therapies \citep{prasad2023rheumatoid}. This wide range of options allows for comparison between RA patients who have continued with TNFi despite other available therapies and those who have tried multiple DMARDs. Previous studies have shown that the majority of patients switched therapies due to loss of efficacy, with 50\% discontinuing their first bDMARD within the first 24 months \citep{strand2017discontinuation}.

Few studies characterized RA patients based on the longitudinal treatment course. Recently, \cite{das2023utilizing} derived treatment sequence clusters under a mixture Markov chain model using EHR data from Mass General Brigham (MGB). It remains unclear how to compare patients across healthcare systems. This study aims to employ a multi-site mixture Markov chain model to cluster RA subjects by their medication sequences using longitudinal EHR data from MGB and Veteran Affairs (VA), assuming that the underlying clusters are shared between MGB and VA but the proportional of patients belonging to the clusters differ across institution. We define treatment clusters based on the longitudinal patterns of how the aformentioned five classes of drugs were used over time for individual patients. 

 We extracted the medication sequences of all eligible people, defined as RA patients who had their first b/tsDMARD
prescription on or after January 1, 2008, with a follow-up period of up to four years. The last column of Table \ref{tab1} shows the sample sizes and basic demographic summary of the two cohorts. We  set the unit of the time window at three months. In cases where a patient takes more than one medication during a time window, the most frequently used medication is selected to represent that period. For periods without medication information, we assume patients adhere to the medication prescribed in the most recent time with records.

%\Xin{add data size, \%patients on each drug for each site. Also most people start from TNFi. Resample the same sample size. Put local model to appendix, add demographic site-specific}

\begin{table}[htbp]
  \centering
\footnotesize
  \caption{Summary statistics for four ensemble clusters in MGB and VA}
  \begin{threeparttable}
    \centering
% Table generated by Excel2LaTeX from sheet 'Sheet1'
    \begin{tabular}{cc|ccccc}
    \toprule
    \multicolumn{2}{c}{} & Cluster 1 & Cluster 2 & Cluster 3 & Cluster 4 & Total \\
    \midrule
    \multirow{2}[1]{*}{N(\%)} & MGB   & 217(10\%) & 292(13.5\%) & 249(11.5\%) & 1406(65\%) & 2164 \\
          & VA    & 267(8.5\%) & 319(10.1\%) & 529(16.8\%) & 2043 (64.7\%) & 3158 \\
              \midrule
    \multirow{2}[1]{*}{Female} & MGB   & 83.4\% & 78.8\% & 85.1\% & 73.3\% & 76.4\% \\
          & VA    & 23.0\% & 21.0\% & 23.3\% & 15.2\% & 17.7\% \\
    \midrule
    \multirow{2}[2]{*}{White} & MGB   & 77.9\% & 77.4\% & 82.7\% & 81.5\% & 80.7\% \\
          & VA    & 81.3\% & 80.2\% & 81.2\% & 78.7\% & 79.5\% \\
    \midrule
    \multirow{2}[2]{*}{Age(sd)} & MGB   & 52.8(15.1) & 56.7(14.9) & 54.5(14.1) & 50.6(15.7) & 52.1(15.5) \\
          & VA    & 61.7(11.0) & 60.0(11.7) & 60.2(11.6) & 60.5(11.5) & 60.5(11.5) \\
    \bottomrule 
    \end{tabular} 
    \begin{tablenotes}
        \item[a] Cluster 1: TNFi to CTLA4; cluster 2: TNFi to Anti-CD20; cluster 3:  multiple b/tsDMARDS; cluster 4: TNFi persisters
    \end{tablenotes}
  \end{threeparttable}
  \label{tab1}
\end{table}%

In this application, the number of study sites $M = 2$ is relatively small to ensure the consistency of the adaptive weights. Therefore, we generate $4$ sub-datasets per site through resampling 80\% of patients with replacement from the original data, thus, creating $M=8$ models for the ensemble learning.
For the $m$-th pseudo-site with $n_m$ resampled patients, we apply a mixture Markov model as in \cite{das2023utilizing}. Specifically, we consider a total of  $S=5$ medication statuses. For the $i$-th patient in the $m$th data set, let $x_{i,t}^{[m]}\in\{1, \dots, S\}$ denote the medication received at time $t$ and $\xx_i^{[m]}\in\{1,\dots,S\}^{h_i}$ the corresponding medication sequence of length $h_i$. We assume there are $K$ latent patient classes and the class membership is denoted as $y_i^{[m]} \in \{1, \dots, K\}$.  Given  $y_i^{[m]} = k$, we model the drug sequence using a Markov chain process with initial state probability $\uu^{[m,k]}\in\R^S$ quantifying the probability of receiving a certain drug as the initial treatment, and transition probability matrix $T^{[m,k]}\in \R^{S\times S}$, where the $(j_1,j_2)$-th entry $T^{[m,k]}$ models the probability of switching from the medication $j_1$ to the medication $j_2$. In this model, the parameters associated with each model are denoted as $\bbeta^{[m]} = (\uu^{[m,1]},\text{vec}(T^{[m,1]}),\cdots,\uu^{[m,K]},\text{vec}(T^{[m,K]}))$. The conditional probability of the observed events and the posterior probability can be expressed as:
%\[
%	P(\xx_i|y_i = l) = \beta_l(\xx_{i,1})B_l(\xx_{i,1}, Y_{i,2})\cdots B_l(\xx_{i,h_{i-1}}, \xx_{i,h_i}).
% \]
\[
P(\xx_i^{[m]}|y_i^{[m]} = l) = \uu^{[m,k]}_{x_{i,1}}T^{[m,k]}_{x_{i,1}, x_{i,2}}\cdots T^{[m,k]}_{x_{i,n_i-1}, x_{i,n_i}},
\]
%\[
%\textstyle P(y_i=l|\xx_i,R_i=m) = \frac{P(y_i=l|R_i=m)P(\xx_i|y_i=l)}  {\sum_{l'=1}^{K}P(y_i=l'|R_i=m)P(\xx_i|y_i=l')}.
%\]
\[
\textstyle P(y_i^{[m]}=l|\xx^{[m]}_i) = \frac{P(y_i^{[m]}=l)P(\xx_i^{[m]}|y_i^{[m]}=l)}  {\sum_{l'=1}^{K}P(y_i^{[m]}=l')P(\xx_i^{[m]}|y_i^{[m]}=l')}.
\]
Within each pseudo-site dataset, $\bbeta^{[m]}$ is estimated using EM algorithm \citep{helske2017mixture}, which will be broadcast to every resampled dataset $m^\prime$ to obtain the corresponding labels $\widehat{\yy}^{[m, m^\prime]}$. After the exchange of parameters, each pseudo-site obtains $M$ sets of predicted labels $\widehat{\YY}^{[m]}=(\widehat{\yy}^{[m, 1]},\cdots,\widehat{\yy}^{[m, M]})\in\{1,...,K\}^{\sum_{m=1}^M n_m}$.  A central analytical server then collects all site-model-specific labels $\{\widehat{\YY}^{[m]}\}_{m=1,...,M}$ and performs ensemble learning. The final clustering results are eventually  returned to each site.

\begin{figure}[H]
    \centering
\includegraphics[width = 6 in]{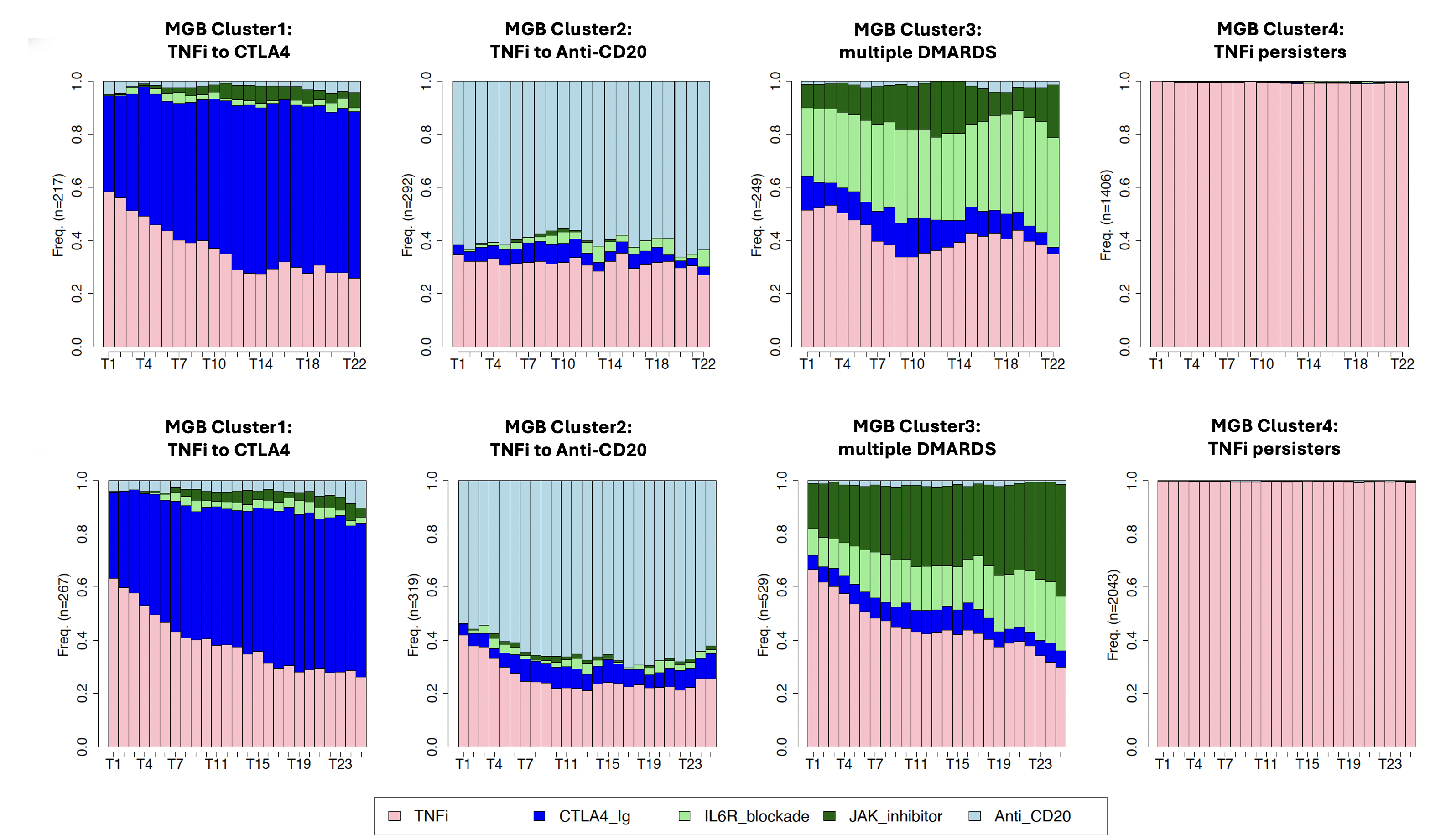}
    \caption{Medication frequency at each time point of the four  clusters identified by FONT. }
\label{fig:ensemble}
\end{figure}

\begin{figure}[H]
    \centering
    \includegraphics[width=1\linewidth, page=2]{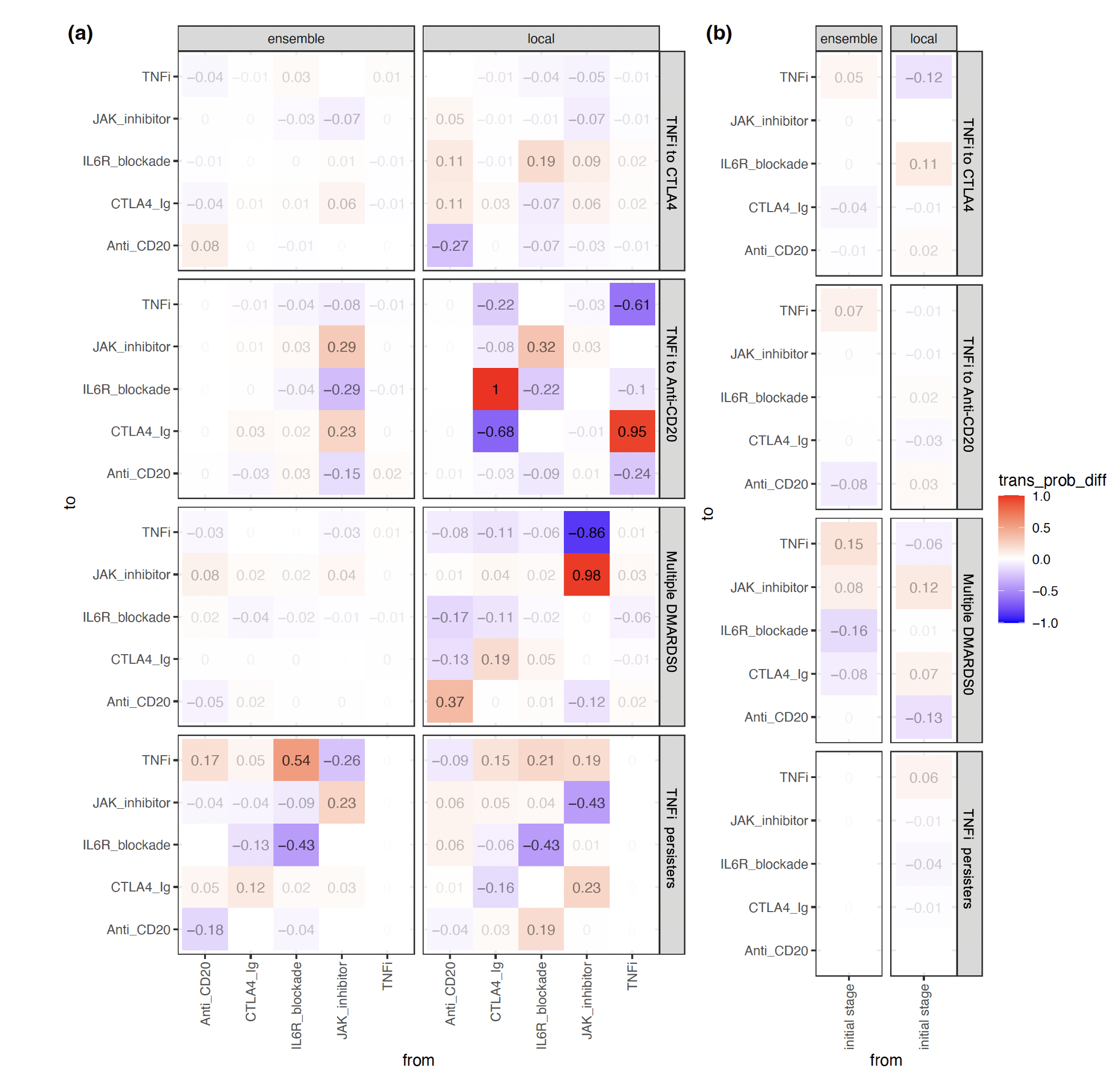}
    \caption{Site difference in (a) transition probability matrices and (b) initial probabilities fitted by Markov models on MGB and VA, stratified by the ensemble (left) or local (right) fitted clustering membership.}
    \label{fig:diff in trans}
\end{figure}

The Bayesian information criterion (BIC) obtained from all local models suggests having $K=4$ best fits our data. Therefore, we choose $K=4$ as the total number of clusters in our analysis. We summarize the site-cluster demographic information in Table \ref{tab1} and visualize the patterns of medication sequences within each cluster by site, as shown in Figures \ref{fig:ensemble}. Based on the observed patterns, cluster 4 consists TNFi persisters who are mostly on TNFi without change of medications. Cluster 1 comprises patients who mostly on CTLA4\_Ig or transiting from TNFi to CTLA4\_Ig. Cluster 2 comprises patients who mostly on anti-CD20 or transiting from TNFi to anti-CD20. Cluster 3 includes patients who are transiting between TNFi, IL6R blockade and JAK inhibitor. Potentially, these patients are less responsive to certain bDMARDs.  The patterns of four clusters are highly consistent when comparing VA and MGB. When the mixture Mardov models are fitted separately on MGB and VA, we observe similar cluster patterns (see Section 2 of Supplementary Materials), but  less aligned cluster characteristics. Specifically, Figure \ref{fig:diff in trans}  compares the empirical estimates of transition matrices and initial probabilities for clusters identified in both MGB and VA. Our analysis shows that clusters identified using FONT are more similar between the two health systems, while clusters identified separately at each site exhibit less similarity. This can be further quantified by the Frobenius norm of the differences in the empirical transition matrices, averaged across clusters (FONT: 0.403 vs. local: 1.083), as well as the differences in initial probabilities (FONT: 0.505 vs. local: 0.515).

The detailed local clustering results obtained from single-site analyses at VA and MGB are presented in Section 2 of the Supplementary Material. Cluster 3 identified by the MGB model has notably smaller sample sizes compared to other clusters. Additionally, the cycling pattern observed in Cluster 3 from the VA model (primarily involving JAK inhibitors alone or transitioning from TNFi to JAK inhibitors) is less transferable to the MGB data. Cluster 1, generated by the local models (VA or MGB), primarily captures the transition between TNFi and CTLA4\_Ig. However, additional medications (e.g., IL6R in Cluster 1 from the VA model, and JAK inhibitors and Anti-CD20 in Cluster 1 from the MGB model) result in less distinguishable clusters. In contrast, the FONT method more clearly differentiates individuals predominantly taking CTLA4\_Ig or shifting from TNFi to CTLA4\_Ig in Cluster 1, and those on IL6R or JAK inhibitors in Cluster 3. This pattern is consistent across both VA and MGB data, with a more pronounced separation between clusters.

%By presenting the difference in the empirical cluster-specific transition matrices and initial probabilities between two sites, Figure \ref{fig:diff in trans} illustrates the corresponding patient clusters in MGB and VA are more similar after applying our ensemble clustering method. 

%We observe that, in general, clusters 2 and 4 identified by different methods are more comparable. Specifically, cluster 2 comprises patients mostly on anti-CD20, while cluster 4 consists mostly of patients on TNFi. Patients in clusters 1 and 3 have been prescribed multiple treatments. Notably, cluster 3 at MGB and VA, as identified by the MGB model, has extremely small sample sizes compared to other clusters, whereas the VA model identified a significantly larger number of samples in cluster 3. Also, the cycling pattern in cluster 3 identified by the VA model (i.e., mainly JAK inhibitor alone, or TNFi to JAK inhibitor) is less transferrable to the MGB data, given the most frequent 30 sequences in MGB cluster 3 in Figure \ref{fig:vamodel} show much disagreement with VA cluster 3. Cluster 1 generated from the local models (VA or MGB) mainly consists of the transition between TNFi and CTLA4\_Ig. However, there exist extra medications (e.g., IL6R in cluster 1 given by VA model, and JAK\_inhibitor and Anti-CD20 in cluster 1 given by MGB models) that complicate the identification of the cluster and lead to less distinguishable clusters with mixed transition schemes. The ensemble model, on the other hand, tends to separate individuals taking CTLA4\_Ig predominantly or shifting from TNFi to CTLA4-Ig in cluster 1 and those on IL6R or JAK inhibitor more in cluster 3. Such a pattern is uniform both in VA and MGB data, while the separation between clusters is more well-marked.

\begin{figure}[H]
    \centering
    \includegraphics[width = 6 in]{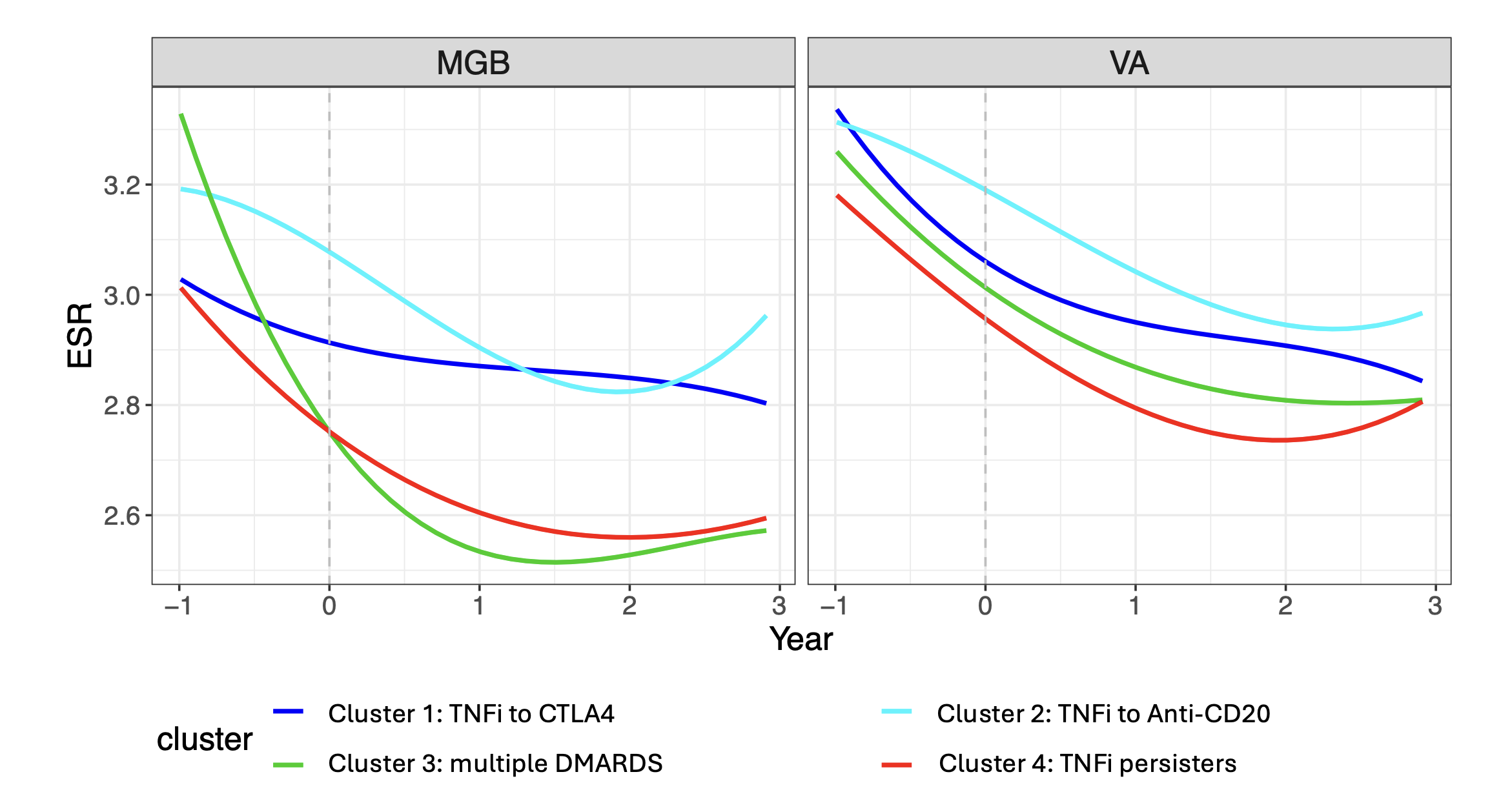}
    \caption{Evaluating the clusters using ESR.}
    \label{fig:labs}
\end{figure}

To further validate the fitted latent classes, we fit a linear mixed effect model with cubic splines for the erythrocyte sedimentation rate (ESR) over time separately in MGB and VA. ESR is commonly used to gauge a person's overall level of inflammation. The cluster membership serves as the fixed effect whereas the random effect occurs at the patient-level. Figure \ref{fig:labs} shows the fitted ESR level for patients in each cluster with time spanning from 1 year before to 3 years after  the first b/tsDMARD prescription. The trend of the ESR over time of each cluster is consistent across the two sites, despite some difference in their actual values. The single-site analyses, as shown in Figure R3 in the Supplementary Material, present less common trends of ESR between two sites, especially in cluster 3. %Specifically, the mean difference in ESR between MGB and VA over time and cluster is 0.167 for the ensemble method, 0.174 for MGB local method, and 0.185 for VA local method. 
To enhance our understanding of the clinical relevance of these clusters, further investigation into the genomic signatures associated with each cluster could provide valuable insights.

\section{Discussion}\label{sec:discussion}
In this paper, we introduce a one-shot federated ensemble learning algorithm for joint clustering analysis across multiple study sites, designed to be compatible with a broad spectrum of existing clustering methods. This method requires sharing only the parameters estimated at each site, along with the cluster memberships predicted for all subjects by each model. 
The approach is expected to be straightforward to implement. Through simulation results across various settings, we demonstrate that our proposed method outperforms several  benchmarks, particularly in scenarios characterized by high site-level heterogeneity and low levels of noise.

Having a large number of models $M$ is crucial for stability of the ensemble. When the number of sites is small, using resampling method to increase the number  of models can improve the performance ({as shown in Section 3.1 of the Supplementary Material}). Our method did not focus on the methods for choosing the number of clusters as it largely depends on the latent class model or the clustering methods to be used to fit the data. In fact, we show in Section 3.2 of the Supplementary Material, overspecifying the number of clusters when fitting  local models has relatively small impact on aggregated distance matrix $\widetilde \bD$. The overall number of clusters can be learned from  $\tilde \D$ based on methods such as the eigenvalue thresholding techniques \citep{ke2023estimation,donoho2023screenot}.

The proposed spectral ensemble learning approach typically assigns higher weights to sites whose clustering outcomes are more consistent with those from other sites. However, incorporating prior knowledge about the reliability of certain sites into the ensemble learning framework could further enhance its effectiveness. This integration would allow for more tailored adjustments using techniques such as constrained eigen-decomposition \citep{gander1989constrained}. Additionally, in cases involving completely irrelevant datasets, such as adversarial scenarios, adapting the method to incorporate sparse assumptions for data-adaptive weights may help mitigate the influence of these datasets, which is worth further investigation. 

\section*{Acknowledgment}
This work was supported by National Institutes of Health (R01GM148494, R01LM013614, R01AR080193, R21AR078339).

\bibliographystyle{unsrtnat}
\bibliography{bibli}

\end{document}